\documentclass[letterpaper]{article} 
\usepackage{aaai2027}  
\usepackage[hyphens]{url}  

\makeatletter
\usepackage[hyphens]{url}
\g@addto@macro\@maketitle{%
  \begin{center}
    \includegraphics[width=\textwidth]{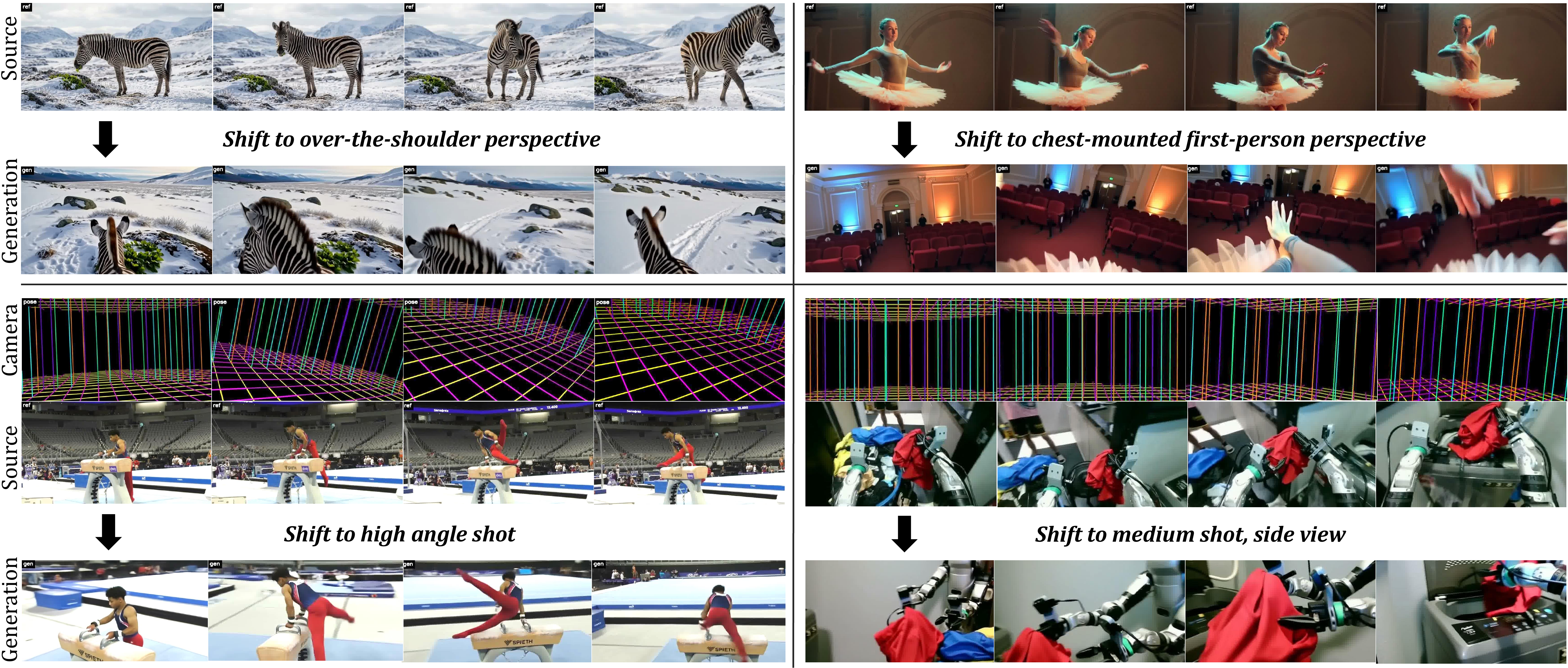}%
    \captionof{figure}{TARS enables robust re-shooting and viewpoint control, plausibly synthesizing unseen regions under large camera motions while supporting reverse-angle shots and seamless perspective switching. \\Project page:  \textcolor{blue}{{\url{https://ymlinfeng.github.io/TARS.github.io/}}}}
    \label{topview} %
  \end{center} %
}

\usepackage{graphicx} 
\usepackage{natbib}  
\usepackage{caption} 
\usepackage{algorithm}
\usepackage{algorithmic}
\usepackage{amsmath}
\usepackage{amssymb}
\usepackage{multirow}

\usepackage{newfloat}
\usepackage{listings}
\DeclareCaptionStyle{ruled}{labelfont=normalfont,labelsep=colon,strut=off} 
\floatstyle{ruled}
\newfloat{listing}{tb}{lst}{}
\floatname{listing}{Listing}

\usepackage{booktabs}

\title{TARS: \underline{T}imestep-\underline{A}ware Data Scaling for 3D-Free Video \underline{R}e-\underline{S}hooting}

\author{
    Jiwen Liu\textsuperscript{\rm 1}\equalcontrib, Shujuan Li\textsuperscript{\rm 1,\rm 2}\equalcontrib, Xiaohan Li\textsuperscript{\rm 1}, Zijie Meng\textsuperscript{\rm 1,\rm 3}, \\ Xinyue Liu\textsuperscript{\rm 1}, Yulong Xu\textsuperscript{\rm 1}, Yan Zhou\textsuperscript{\rm 1}, Guoxin Zhang\textsuperscript{\rm 1}
}
\affiliations{
    \textsuperscript{\rm 1}Kling Team, 
    \textsuperscript{\rm 2}Tsinghua University,
    \textsuperscript{\rm 3}Peking University\\
}

\begin{document}

\maketitle 
\begin{abstract}
 
Video re-shooting aims to regenerate videos with controllable camera motion and viewpoint. Existing methods rely on explicit 3D priors, which are limited by reconstruction quality and often perform poorly when synthesizing previously unseen regions, or on paired videos with different camera trajectories, whose scarcity hinders generalization. 
We revisit video re-shooting through text-driven semantic viewpoint specification, enabling control over shot scale, viewing angle, and first-/third-person perspective. To this end, we propose TARS, a 3D-free video re-shooting paradigm. Timestep-wise sensitivity analysis reveals that camera motion is primarily established during high-noise stages, where coarse spatiotemporal structures are formed. Based on this insight, we introduce self-supervised training to learn camera dynamics and fundamental visual representations without paired re-shooting data or 3D reconstruction. Through data scaling and joint textual-camera conditioning, TARS supports robust camera and viewpoint control, plausibly synthesizing regions beyond the source view under large camera motions while enabling reverse-angle re-shooting and perspective switching.
Extensive experiments show that TARS provides more accurate and temporally consistent camera control than prior methods.
\end{abstract}

\begin{figure}[t]
\begin{center}
\includegraphics[width=0.95\linewidth]{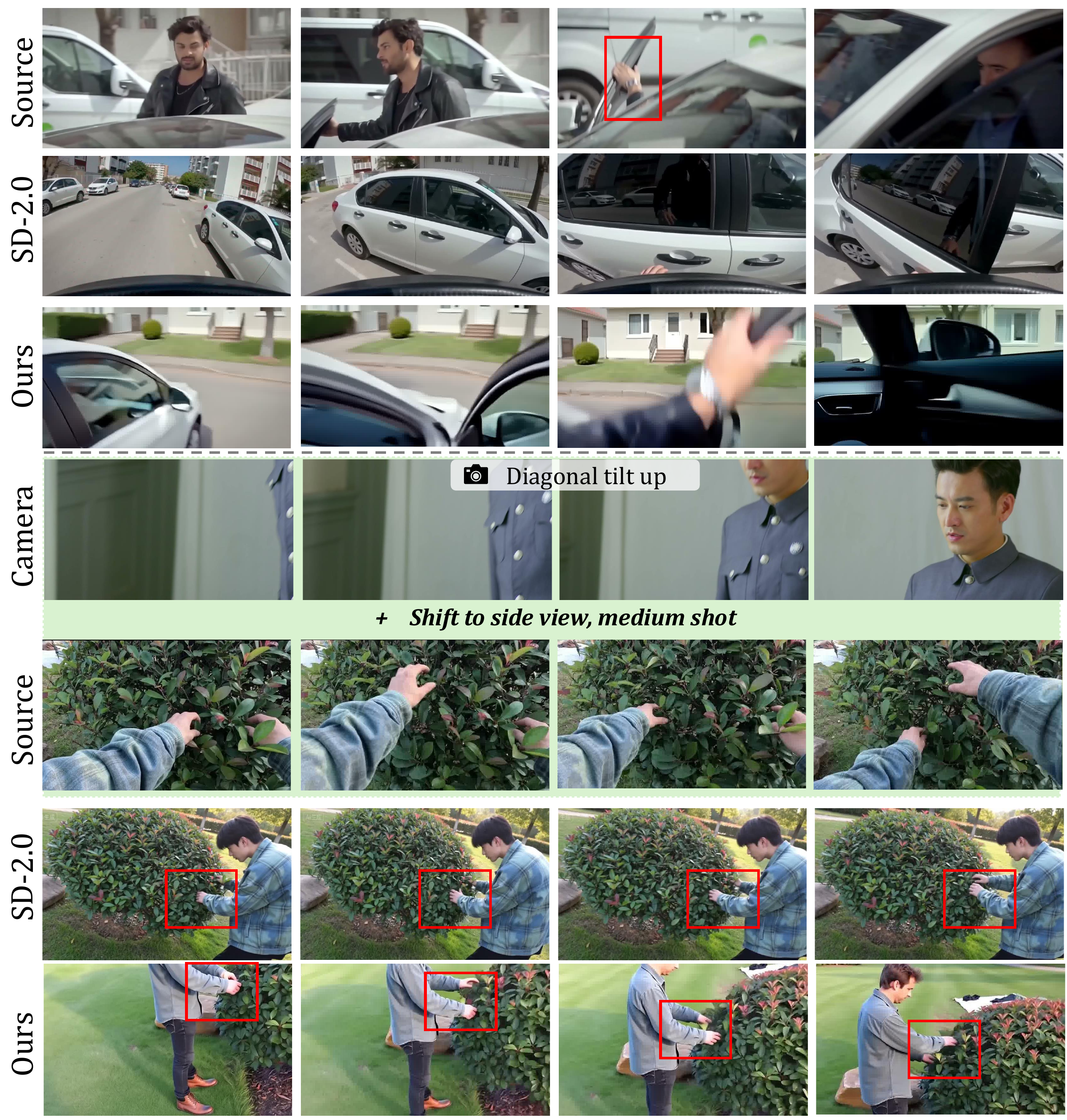}
\end{center}
\caption{\textbf{Comparison with SD-2.0 in viewpoint control}. Our method enables accurate transition between the third-person view and the first-person view.}
\label{fig:vis_fov}
\vspace{-0.5cm}
\end{figure}

Video re-shooting~\cite{lin2026vista4d, bai2025recammaster, luo2025camclonemaster, ma2025controllable} aims to regenerate a source video under controlled camera motion and viewpoint while preserving its subjects, actions, and scene content. It has broad practical applications in cinematography, advertising, autonomous driving, and embodied artificial intelligence. However, as a source video provides only a single-view observation at each timestamp, large camera movements may expose occluded or previously unobserved regions. The model must therefore plausibly synthesize these regions while maintaining spatial and temporal consistency with the source video, making video re-shooting particularly challenging.

Existing video re-shooting methods broadly follow two paradigms: leveraging explicit 3D/4D priors or learning directly from paired videos that capture the same scene with different camera trajectories. Prior-based methods~\cite{lin2026vista4d, yu2025trajectorycrafter, Ren_2025_CVPR} typically reconstruct the source video into explicit 3D/4D representations~\cite{Wang_2025_CVPR, lin2025dpa3, ICLR2025_ce1710ff, Li_2025_CVPR} and render them along the target camera trajectory. Besides introducing substantial computational overhead, their performance is limited by reconstruction quality, making them vulnerable to accumulated geometric errors and less effective at synthesizing regions outside the source view. Alternatively, end-to-end approaches~\cite{bai2025recammaster, luo2025camclonemaster} learn camera transformations from such paired videos. However, high-quality paired videos are difficult to collect, often leading to overfitting and limited generalization to diverse scenes and large camera motions. Moreover, existing formulations typically either constrain the target viewpoint to that of the source video or require hard-to-obtain camera parameters to redefine it, offering limited semantic control over shot scale, viewing angle, and perspective.

To address these limitations, we propose TARS, a 3D-free training paradigm for video re-shooting that enables flexible camera and viewpoint control with minimal paired data. Beyond conventional geometric control, TARS introduces text-driven semantic viewpoint specification, allowing users to describe the desired shot scale, viewing angle, and perspective without providing explicit camera parameters. By combining textual instructions with camera-motion conditions, TARS controls both how the scene is viewed and how the camera moves over time, supporting flexible semantic and geometric re-shooting as shown in Figure~\ref{fig:vis_fov}.

The design of TARS is motivated by a timestep-wise sensitivity analysis of video diffusion models. We find that high-noise denoising stages play a dominant role in establishing coarse spatiotemporal structures, particularly camera motion and subject dynamics, whereas later stages mainly refine appearance and visual details. Based on this observation, we first construct large-scale self-supervised data to jointly learn camera motion, viewpoint transformations, and fundamental visual representations. Scaling up such data substantially improves viewpoint controllability and generalization, enabling challenging transformations such as large camera movements, reverse-angle re-shooting, and switching between first- and third-person perspectives. Finally, rather than relying heavily on paired re-shooting videos, we route only a small amount of cross-pair data to the high-noise stages, which effectively elicits temporally synchronized subject motion under novel camera transformations. With these designs, TARS can plausibly synthesize regions beyond the source view while preserving the original content and motion. Extensive experiments demonstrate that TARS achieves more accurate and temporally consistent camera control than state-of-the-art methods and generalizes well across diverse scenes, viewpoints, and camera motions.
\begin{figure}
    \centering
    \includegraphics[width=0.95\linewidth]{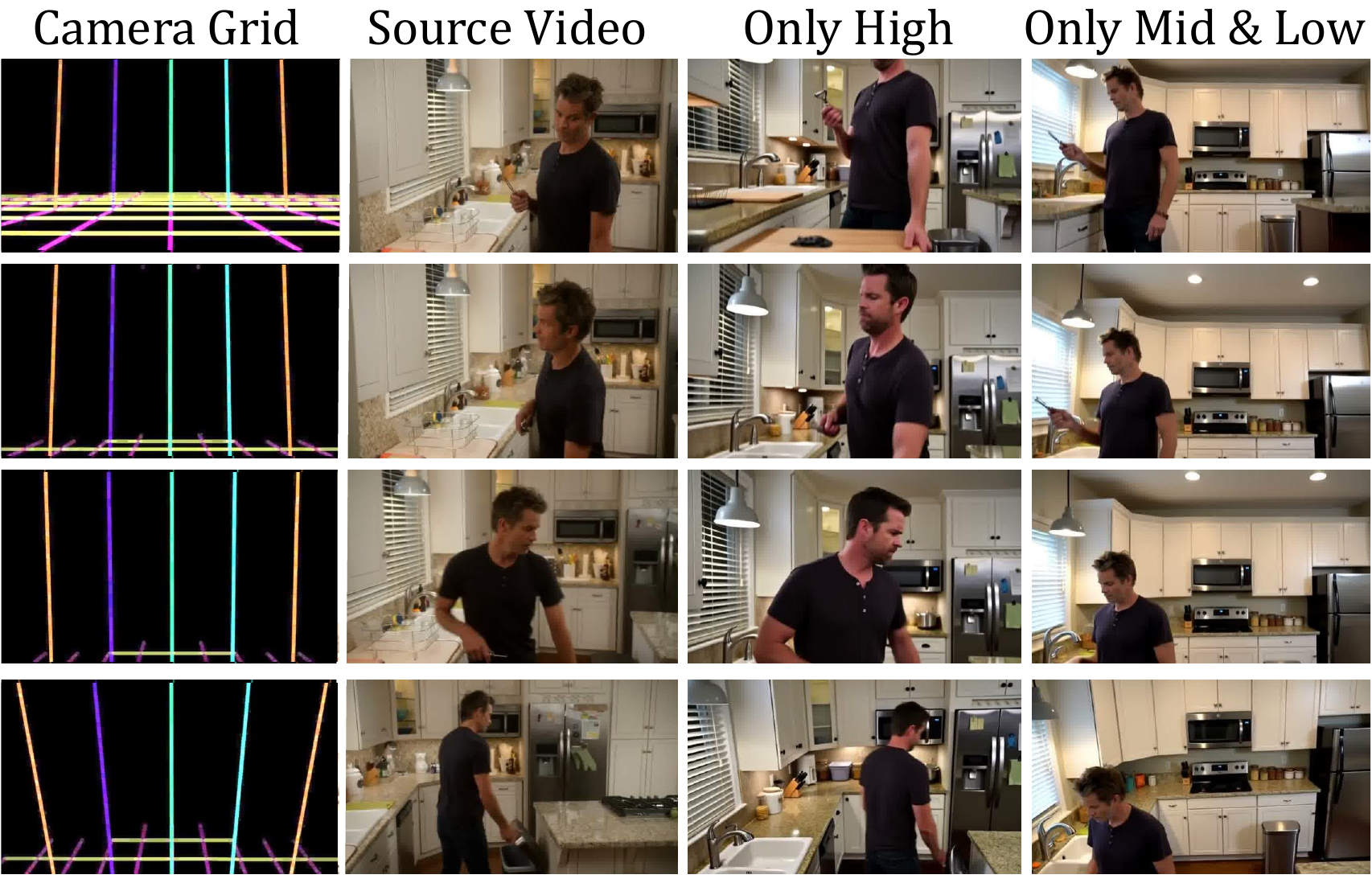}
    \caption{\textbf{Visualization of source video injection at different diffusion timesteps}. High-noise timesteps capture low-frequency structure and motion, while rest timesteps preserve high-frequency details such as subject identity.}
    \label{fig:timestep_analysis}
    \vspace{-0.5cm}
\end{figure}
In summary, our main contributions are:
\begin{itemize}
    \item We introduce the first video re-shooting framework that unifies camera re-shooting and text-driven semantic viewpoint control, enabling accurate camera trajectory manipulation together with flexible control over shot scale, viewing angle, and first-/third-person perspective. 
    
    \item We reveal the timestep-dependent roles of video diffusion models and propose a 3D-free training strategy that learns camera motion and viewpoint transformations from large-scale self-supervised data, while using minimal cross-pair data to ensure motion synchronization.
    
    \item Extensive experiments demonstrate superior control accuracy, temporal consistency, and generalization, particularly for large camera motions and semantic viewpoint transformations.
\end{itemize}

\section{Related Work}
\begin{figure*}[t]
    \centering
    \includegraphics[width=0.95\linewidth]{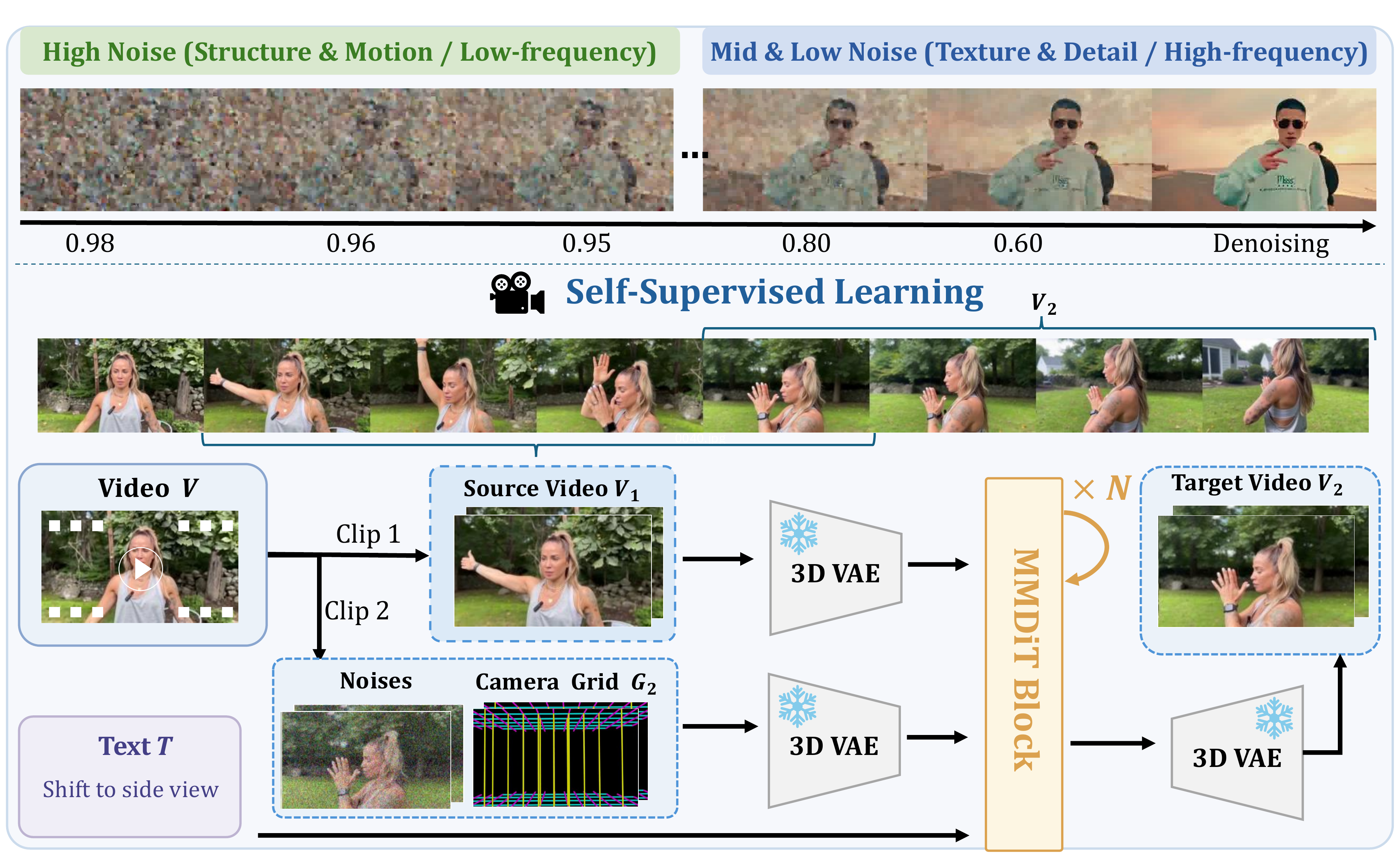}
    \caption{\textbf{Timestep-aware self-supervised learning framework}. \textbf{Top:} High-noise diffusion timesteps primarily learn global structure, viewpoint, and camera motion, while mid- and low-noise timesteps refine texture details. \textbf{Bottom:} Guided by this observation, we construct self-supervised training pairs by temporally splitting each video into two clips, enabling large-scale learning of camera transitions from unlabeled videos.}
    \label{fig:self-learning}
\end{figure*}
\subsection{Video Diffusion Models}
Propelled by Diffusion Models~\cite{ho2020denoising, ho2022video, ho2022imagen}, video generation has rapidly evolved across various modalities~\cite{singer2022make, villegas2023phenaki, polyak2024movie, bar2024lumiere, xing2024dynamicrafter, chen2023seine, renconsisti2v, wang2023videocomposer,chen2024videocrafter2} (e.g., Text/Image/Video-to-Video) and multi-signal fusion paradigms~\cite{singer2022make, zhang2023i2vgen, wang2023videocomposer, kong2024hunyuanvideo}. Crucially, this rapid advancement is heavily underpinned by scaling laws and massive datasets, which is essential for learning robust spatio-temporal priors and complex open-world dynamics.

\subsection{Camera Control in Video Generation}
Although recent controllable video generation models~\cite{blattmann2023stable, luo2026shotstream, guo2023animatediff, xu2024camco, he2024cameractrl, wang2024motionctrl, zheng2024cami2v,  meng2026make, hu2024motionmaster, he2025cameractrlii, luo2026shotstream, wang2025cinemaster} have achieved remarkable progress in camera control by incorporating pose conditions, camera trajectory re-shooting~\cite{bai2025recammaster, lin2026vista4d} for existing videos remains a challenging task.

Existing re-shooting methods can be broadly categorized into two paradigms. The first paradigm~\cite{lin2026vista4d, Ren_2025_CVPR, wang2025cinemaster, yu2025trajectorycrafter} relies on explicit 3D or 4D scene representations~\cite{Wang_2025_CVPR, lin2025dpa3}, which perform spatio-temporal reconstruction of the source video followed by novel view rendering. 

While these methods perform well in static or subtle-motion scenes, they are highly susceptible to severe artifacts when confronted with highly dynamic scenes, complex occlusions, or large-scale viewpoint transformations, hindering their generalization to complex open-domain videos.

The second paradigm~\cite{bai2025recammaster, luo2025camclonemaster, vanhoorick2024gcd, ICLR2025_VD3D} attempts to directly train generative models in an end-to-end manner using cross-pair data, aiming to learn the mapping from the source video to the target camera trajectory. 
However, these methods suffer from severe data starvation, which prevents the model from adequately decoupling camera motion from subject dynamics. This often leads to spatio-temporal ambiguity during training, ultimately failing to faithfully preserve the motion consistency of the subjects during re-shooting.

\subsection{Timestep-Aware Diffusion}
The iterative denoising process in diffusion models inherently operates in a coarse-to-fine manner. Pioneering works like eDiff-I ~\cite{balaji2022ediff} demonstrate that high-noise stages predominantly construct global structures, while low-noise stages focus on synthesizing high-frequency details. Similarly, frequency-domain analyses such as FreeU~\cite{si2024freeu} reveal that high-noise steps dictate global spatial layouts, whereas low-noise steps are dedicated to local texture refinement.
These timestep-aware properties provide a crucial foundation for our method. Since camera motion is a global, low-frequency geometric transformation, forcing the model to learn it uniformly across all noise levels is highly data-inefficient and risks compromising fine details.

\section{Method}
\subsection{Preliminary}
\subsubsection{Camera Grid} is a video-format representation~\cite{liu2025omnidirector} of camera motion that encodes camera parameters as visual grid transformations within an empty 3D room. 
Given its universality and ease of injection into diffusion models, we adopt this camera representation in our method.

Specifically, given a sequence of camera trajectory parameters $P_i = [R_i | t_i]$, where $R_i$ denotes the rotation matrix and $t_i$ represents the translation vector, the camera grid first computes the spatial bounding box of the trajectory. Subsequently, it samples two X-Z planes at distinct heights, $y_{\text{ceiling}}$ and $y_{\text{floor}}$, to represent the ceiling and the floor, respectively. Grid points $\{\mathbf{p}\}$ are then uniformly sampled on these two planes, establishing a fundamental spatial reference frame. 

As the camera moves along the trajectory, the camera grid renders the spatial grid points via perspective projection with the camera extrinsic $P_i$ and intrinsics $K$, formulated as:
\begin{equation}
  \lambda \mathbf{u} = K (R_i \mathbf{p} + t_i)  
\end{equation}
where $\mathbf{p} = [x, y, z]^T$ is the 3D coordinate of a sampled grid point, $\mathbf{u} = [u, v, 1]^T$ represents the projected 2D homogeneous coordinate on the image plane, and $\lambda$ denotes the depth of the point relative to the camera. 

By connecting the projected points in the image plane to form intersecting lines on the horizontal planes as well as vertical lines bridging the two planes, the camera grid explicitly translates variations in camera parameters into the dynamic movement of a spatial grid. Pairing the camera grid with the video enables self-supervised camera motion injection ~\cite{liu2025omnidirector}.
 
\subsubsection{Rectified Flow.} 
The Rectified Flow framework~\cite{esser2024scaling} mitigates the randomness in the generation process by formulating the denoising procedure of diffusion models as a deterministic probability flow. Its core idea is to connect the empirical data distribution and the standard normal distribution via straight-line paths. The forward process is defined as a linear interpolation between the clean data and the noise, formulated as:
\begin{equation}
    z_t = (1 - t) z_0 + t \epsilon
\end{equation}
where $ \epsilon \sim \mathcal{N}(0, \mathbf{I}) $ denotes the standard Gaussian noise, $ z_0 $ represents the clean data, and $ z_t $ is the latent state at continuous time $ t \in [0, 1] $. The derivative of this forward process, which defines the ordinary differential equation (ODE), is given by:
\begin{equation}
\frac{\mathrm{d}z_t}{\mathrm{d}t} = \epsilon - z_0    
\end{equation}

During training, a neural network parameterized by \( \theta \) is employed to estimate the vector field \( v_\theta \). The model is optimized via conditional flow matching to learn this straight-line probability path. The training objective~\cite{lipman2023flow} is defined as:
\begin{equation}
    \mathcal{L} = \mathbb{E}_{t \sim \mathcal{U}(0,1), z_0, \epsilon} \left[ \left\| v_\theta(z_t, t \mid c) - (\epsilon - z_0) \right\|^2 \right]
\label{loss_flow}
\end{equation}

where $ c $ denotes the injected conditioning signals. During inference, data generation is achieved by solving this ODE backward from $t = 1 $ to $ t = 0 $.

The objective of the video re-shooting task is to generate a target video \( V_{tgt} \) that follows a specified target camera trajectory \( P' \), given a source video \( V_{src} \) and auxiliary control signals such as text prompts \( T \) for viewpoints. Under this setting, \( V_{src} \) and \( V_{tgt} \) differ exclusively in their camera motion and viewpoints, while strictly sharing identical visual appearances and temporal action dynamics. In our implementation, we employ a camera grid \( G \) as the explicit representation of the target camera motion \( P' \). 
As defined in Equation~\ref{loss_flow}, we instantiate the target data \( z_0 \) as the latent representation of the target video \( V_{tgt} \), and formulate the conditioning signal as a comprehensive set \( c = \{V_{src}, G, T\} \). By doing so, the network \( v_\theta \) is trained to predict the vector field that flows towards the target video latents, strictly guided by the source video, the desired camera grid, and the text prompt.
\begin{figure*}[t]
\begin{center}
\includegraphics[width=0.99\linewidth]{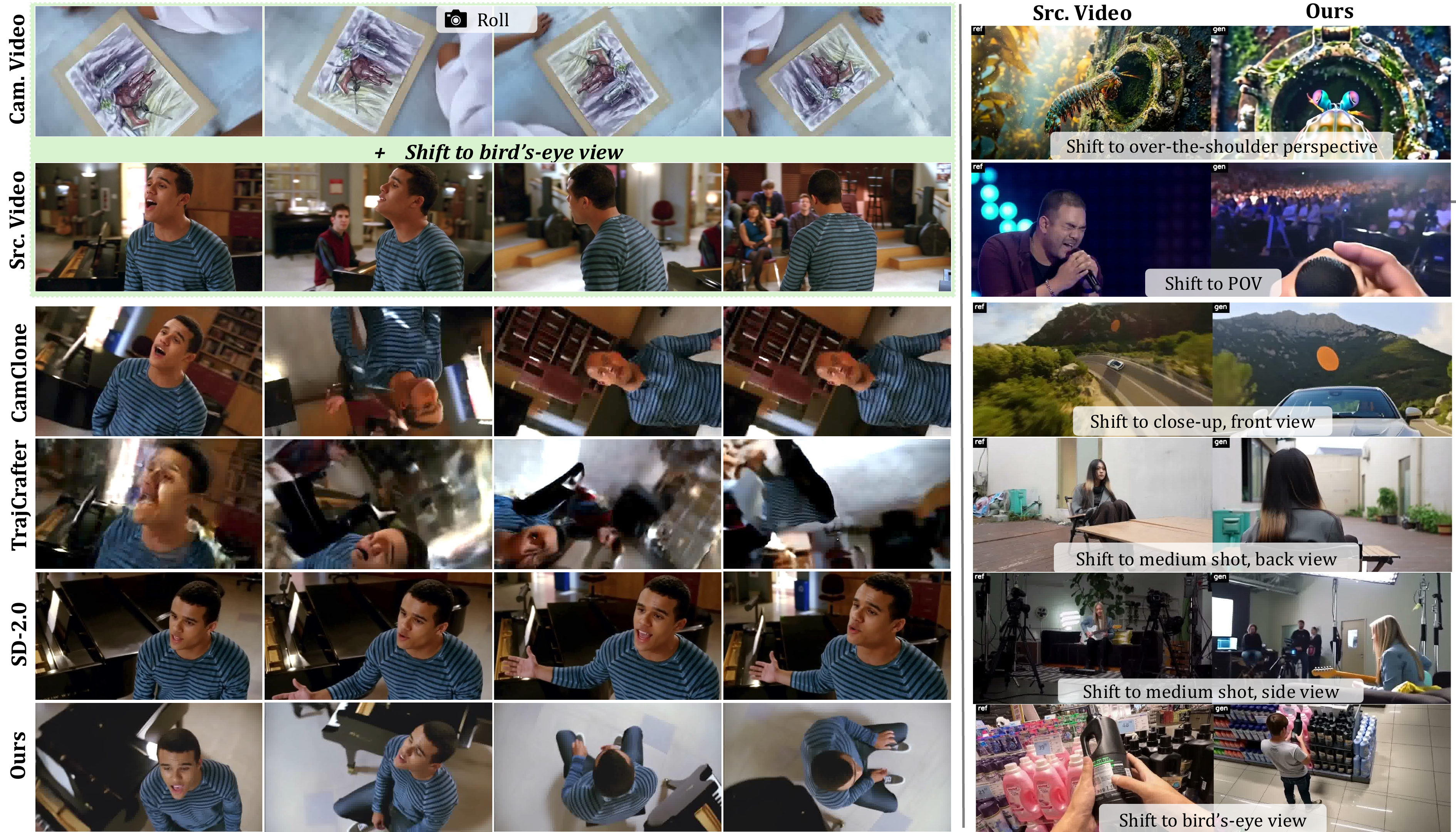}
\end{center}
\caption{\textbf{Qualitative Evaluations.} The results demonstrate that TARS can accurately reshoot the source video under novel camera trajectories and viewpoints.}
\label{fig:qualitive}
\vspace{-0.5cm}
\end{figure*} 
\subsection{Analysis of Learning Dynamics Across Timesteps}
During the denoising process, the model exhibits significant non-uniformity in generating information across different frequencies at varying timesteps. A similar frequency preference has been observed in recent studies on image diffusion models~\cite{balaji2022ediff, si2024freeu}. We visualize the inference process based on the re-shooting model trained on the cross-paired data to analyze the phenomenon. 

As demonstrated in the top of Figure~\ref{fig:self-learning}, we observe that the model rapidly establishes the global spatial structure and motion dynamics within a very small fraction of early iteration steps, while dedicating the subsequent extensive iterations to refining local appearances. This indicates that the model primarily learns low-frequency signals (such as camera and motion) in the high-noise regime, and focuses on high-frequency texture details in the mid-to-low-noise regime.

We further validate this observation by injecting the source video $V_{\mathrm{src}}$ at different diffusion timesteps. Specifically, we first condition the model on $V_{\mathrm{src}}$ only during the high-noise stage and remove this signal for all subsequent timesteps. As shown in the ``Only High'' column of Figure~\ref{fig:timestep_analysis}, the generated video largely preserves the spatial layout and human motion of the source video, while exhibiting substantial drift in subject identity. In contrast, when $V_{\mathrm{src}}$ is injected only at the mid-to-low-noise timesteps (``Only Mid$\&$Low''), the subject identity remains consistent with the source video, whereas the motion patterns and spatial layout are largely lost. These results further demonstrate that diffusion models primarily capture low-frequency information, such as global scene structure and motion, at high-noise timesteps, while recovering high-frequency appearance details, including subject identity and texture, during the mid-to-low-noise stages.

In real-world data collection, strictly paired data capturing the same scene with different camera trajectories (i.e., cross-pair data) is extremely scarce, whereas data sharing identical high-frequency information (e.g., appearance and texture) is relatively abundant. However, existing methods typically rely on such scarce paired data across all timesteps during training. In fact, for the vast majority (potentially up to 95\%) of the denoising steps, such strict pairing constraints are unnecessary. Therefore, designing a timestep-aware data strategy can significantly enhance data utilization efficiency and alleviate the bottleneck caused by the scarcity of real-world cross-pair data. Motivated by this analysis, we introduce our proposed Timestep-Aware Data Routing strategy in the following section.
\subsection{Timestep-Aware Data Routing}
To enhance data utilization efficiency, we construct differentiated paired data \( \{V_{src}, V_{tgt}\} \) tailored to different timesteps. First, we rewrite the objective in Eq.~\ref{loss_flow} for video re-shooting as follows:
\begin{equation}
    \mathcal{L} = \mathbb{E} \left[ \| v_\theta(z_t, t \mid V_{src}, G_{tgt}, T) - (\epsilon - z_0) \|_2^2 \right]
\label{reshoot_loss}
\end{equation}
where \( z_0 = \text{VAE}(V_{tgt}) \), and the camera grid \( G_{tgt} \) shares the identical camera trajectory with the target video \( V_{tgt} \). For simplicity, the subscript of the expectation symbol is omitted. 
Specifically, we propose a timestep-aware two-stage data strategy: the first stage learns general video priors through large-scale self-supervised learning, while the second stage focuses on motion alignment by introducing limited paired supervision only in the high-noise regime.

\subsubsection{Stage 1: Self-Supervised Learning for Re-Shooting.} Given that constructing the target camera grid \( G_{tgt} \) from \( V_{tgt} \) is straightforward, and that the majority of noise steps focus on appearance generation, we initially disregard the motion consistency and concentrate on camera trajectories and appearance. Specifically, we randomly sample a complete video \( V \) and split it into two video clips of equal duration, denoted as \( V_1, V_2 \in \mathbb{R}^{3 \times F \times H \times W} \), where \( F \) represents the number of frames. Subsequently, we estimate the camera poses of \( V_2\) and render the corresponding camera grid following \( G_2 \)~\cite{liu2025omnidirector}. 

To further introduce explicit viewpoint semantics, we employ a multimodal large language model (e.g., Qwen3-VL~\cite{bai2025qwen3}) to generate textual descriptions of the viewpoint transition. Specifically, we prompt the MLLM to decompose the camera transition into three orthogonal attributes:  shot scale (close-up, medium shot, and long shot), viewing angle (front, side, back, high-angle, and low-angle), and perspective (over-the-shoulder, first- or third-person). Such attribute-level descriptions provide a structured semantic representation of viewpoint, allowing the diffusion model to learn fine-grained associations between textual instructions and viewpoint changes. The resulting description serves as the text condition $T$, and
the learning objective for this stage can thus be formulated as:
\begin{equation}
\mathcal{L}_{self} = \mathbb{E} \left[ \| v_\theta(z_t, t \mid V_1, G_2, T) - (\epsilon - z_2) \|_2^2 \right]
\end{equation}
where \( z_2 = \text{VAE}(V_2) \). By optimizing this objective, we achieve the preliminary injection of structure, viewpoint, camera motion, and appearance. Since the paired data are constructed from the same source video, they inherently maintain strict consistency in high-frequency information. Moreover, we can flexibly adopt overlapping or non-overlapping sampling strategies, where large-gap non-overlapping sampling substantially improves the model's ability to hallucinate unseen regions.
\begin{table*}[t]
\centering
\begin{tabular}{l|cc|ccccc|ccc}
    \toprule
    \small
    \multirow{2}{*}{Method} & \multicolumn{2}{c|}{\textbf{Camera Accuracy}} & \multicolumn{5}{c|}{\textbf{Spatio-temporal Consistency}} & \multicolumn{3}{c}{\textbf{Visual Quality}} \\
    \cmidrule{2-11}
    & R-Pre $\%$ $\uparrow$ & T-Pre $\%$ $\uparrow$ & MPGE $\downarrow$ & Arc$\uparrow$ & GCR $\downarrow$ &LSR $\uparrow$& FSCS$\%$ $\uparrow$&  CE $\uparrow$ &  FDR$\%$ $\downarrow$& VDR$\%$ $\downarrow$ \\
    \midrule
    Camclone
    & \underline{71.77} & \underline{51.48} 
    &\underline{21.67}$^\circ$  &0.313 &1.192&0.652&71.07
    &0.57&17.82&61.49\\
    TrajCrafter
    & 67.24 & 47.21
    &31.38$^\circ$   &0.180 &1.432&0.473&47.33
    &0.14&39.10&91.22 \\
    SD-2.0
    & 57.23 & 28.91
    &27.77$^\circ$  &\textbf{0.43}  &\textbf{0.901}&\textbf{0.723} &\underline{74.15}
    &\underline{0.89}&\textbf{2.99}&\textbf{11.15} \\
    Ours
    & \textbf{83.18}& \textbf{72.76}
    &\textbf{10.34$^\circ$ }&\underline{0.41}  &\underline{0.916}&\underline{0.709} &\textbf{83.92}
    &\textbf{0.91}&\underline{3.48}&\underline{12.57}\\
    \bottomrule
\end{tabular}
\caption{\textbf{Quantitative Evaluations.} In comparison, we achieve superior performance across most evaluation metrics, which demonstrates the effectiveness of TARS. }
\vspace{-0.5cm}
\label{recam_comparison}
\end{table*}
\begin{table}
    \centering
    \small
    \begin{tabular}{l|c|c|c}
    \toprule
         Method & Shot Scale & Viewing Angle & Perspective   \\
    \midrule
    SD-2.0 &  0.69& 0.72 & 0.47\\
    Ours     & \textbf{0.83} & \textbf{0.81} & \textbf{0.98} \\
    \bottomrule
    \end{tabular}
    \caption{Quantitative comparison with SD-2.0~\cite{seedance2026seedance} on \textbf{viewpoint accuracy.}}
    \label{tab:viewpoint_comparison}
    \vspace{-0.5cm}
\end{table}

\subsubsection{Stage 2: Eliciting Spatio-Temporal Consistency via Few Cross-Paired Data.} Although Stage 1 achieves ``semantic'' video re-shooting  
where the camera trajectory aligns with the target video and the appearance resembles the source video, it still falls short in temporal dynamics. Because we disregarded spatio-temporal consistency (i.e., subject motion in the source video) during data construction, the current model only captures semantic-level coherence. For instance, if the complete video \( V \) depicts a person walking forward, $ V_1$ and $V_2 $ represent the first and second halves of the walking sequence, respectively, which are not temporally synchronized.
To address this limitation, we utilize a small amount of cross-pair data to perform a second-stage fine-tuning only in the high-noise regime according to Equation~\ref{reshoot_loss}, while the remaining diffusion timesteps continue to be trained with the self-supervised data. This design injects spatio-temporally aligned motion supervision, enabling temporally consistent video re-shooting. Meanwhile, since the majority of diffusion timesteps remain optimized using large-scale self-supervised data, the learned data distribution is largely preserved, preventing the model from overfitting to the limited synthetic cross-pair data and maintaining its strong generalization. 

Fundamentally, synthesizing \( V_2 \) conditioned solely on \( V_1 \) is an inherently one-to-many problem with multiple plausible motion patterns. In experiments, we observe a \emph{shortcut effect}: when multiple valid solutions exist, the diffusion model naturally tends to converge to the solution whose motion is temporally synchronized with the input video, as this solution is easier to optimize. This tendency is further reinforced by text conditioning and classifier-free guidance (CFG), which bias the denoising process toward this synchronized mode. Therefore, even a small amount of strictly paired supervision is sufficient to activate this behavior, encouraging the model to exploit the motion cues in \( V_{src} \) and thereby achieving accurate spatio-temporal synchronization.

\section{Experiments}
\subsection{Experimental Setups}
TARS is initialized from an in-house text-to-video diffusion model. For the training data, we curate 1M video clips for the first-stage self-supervised training. In the second stage, we utilize 60K cross-pair samples, comprising 10K high-quality real-world videos with rich subject motions and camera movements, alongside 50K Unreal Engine (UE) videos featuring simple camera trajectories. We train the model for 8K iterations with a learning rate of \(5 \times 10^{-5}\). We empirically define the high-noise regime as the timesteps \( t \in [0.95, 1.0] \). For cases where the source video is in the third-person perspective while the target video requires a first-person perspective, we discard the camera trajectory and instead allow the camera to follow the motion of the first-person subject.

\subsubsection{Evaluation Metrics.}
We evaluate the performance of TARS across four dimensions:(1) \textbf{Camera Accuracy}: we report R-Prec and T-Prec estimated by DPA-V3~\cite{lin2025dpa3}. (2)\textbf{Viewpoint Accuracy:} we evaluate viewing angle, shot scale, and perspective using Gemini 3.1 Pro, and compare with Seedance 2.0~\cite{seedance2026seedance}. (3) \textbf{Spatio-temporal Synchronization} is evaluated on motion (V-MPGE), identity (ArcFace~\cite{Deng_2019_CVPR}), and scene consistency (GCR and LSR).
(4) \textbf{Visual Quality}: We evaluate content expansion (CE) and visual distortion using Frame-level Distortion Rate (FDR) and Video-level Distortion Rate (VDR), where all metrics are assessed by Gemini 3.1 Pro.

\begin{table*}
\centering
\begin{tabular}{l|cc|ccccc|ccc}
    \toprule
    \multirow{2}{*}{Method} & \multicolumn{2}{c|}{\textbf{Camera Accuracy}} & \multicolumn{5}{c|}{\textbf{Spatio-temporal Consistency}} & \multicolumn{3}{c}{\textbf{Visual Quality}} \\
    \cmidrule{2-11}
    & R-Pre $\%$ $\uparrow$ & T-Pre $\%$ $\uparrow$ & V-MPGE $\downarrow$ & ArcFace $\uparrow$ & GCR $\downarrow$ &LSR $\uparrow$& FSCS $\uparrow$&  CE $\uparrow$ &  FDR $\downarrow$& VDR $\downarrow$ \\
    \midrule
    Mid\&Low&
    75.44&68.82&
    23.54$^\circ$&0.38&1.205&0.598&41.09&
    0.54&9.74&23.95
    \\
    High&
    82.14&71.34&
    12.33$^\circ$&0.28&1.234&0.612&71.62&
    0.81&8.68&21.34
    \\
    Stage1&
    81.91&67.75&
    18.23$^\circ$&0.39&0.992&0.673&74.78&
    0.85&4.12&16.39
    \\
    Stage2&
    69.47&48.12&
    21.27$^\circ$&0.36&1.164&0.617&68.09&
    0.62&13.98&26.36
    \\
    Full& 
    \textbf{83.18}& \textbf{72.76}
    &\textbf{10.34$^\circ$}&\textbf{0.41}  &\textbf{0.916}&\textbf{0.709} &\textbf{83.92}
    &\textbf{0.91}&\textbf{3.48}&\textbf{12.57}
    \\
    \bottomrule
    
\end{tabular}
\caption{Results of the ablation study for different experimental settings.}
\label{abaltion_camera}
\end{table*}

\begin{table}
    \centering
    \small
    \begin{tabular}{l|c|c|c}
    \toprule
         Method & Shot Scale & Viewing Angle & Perspective   \\
    \midrule
    Mid\&Low& 0.75& 0.78&0.89
    \\
    High&  0.81& 0.79& 0.97
    \\
    Stage1& 0.79&  0.78& 0.96
    \\
    Stage2& 0.71& 0.69&  0.36
    \\
    Full& \textbf{0.83}&\textbf{0.81}&\textbf{0.98}
    \\
    \bottomrule
    \end{tabular}
    \caption{Ablation studies on \textbf{viewpoint accuracy.}}
    \label{tab:viewpoint_ablation}
    \vspace{-0.5cm}
\end{table}
  
\subsubsection{Evaluation Set.}
We curate a diverse and challenging validation set consisting of 1021 samples. All samples are collected from online videos and cover diverse scenarios, including humans, objects, animals and landscapes. We categorize the videos according to two dimensions, camera motion complexity and scene dynamics, each of which is divided into simple and complex levels, resulting in four categories. We then sample approximately the same number of examples from each category to construct a balanced evaluation set.

\subsection{Comparisons with State-of-the-Art Methods}
\subsubsection{Baselines.} We compare TARS with state-of-the-art re-shooting methods, primarily focusing on the 3D-based approach, TrajCrafter~\cite{yu2025trajectorycrafter}, the cross-pair-based method, CamClone~\cite{luo2025camclonemaster} and commercial model SD-2.0~\cite{seedance2026seedance}. Specifically, TrajCrafter~\cite{yu2025trajectorycrafter} first reconstructs 3D point clouds and projects them with the target camera trajectory for video generation. In contrast, CamClone~\cite{luo2025camclonemaster} trains a Diffusion Transformer (DiT) using cross-pair data. 

\subsubsection{Qualitative comparison.}

The visual comparisons with baseline methods are provided in Figure~\ref{fig:qualitive}. CamClone~\cite{luo2025camclonemaster} demonstrates strong camera-following capability, but often suffers from severe texture collapse and unrealistic appearance. We attribute this limitation to its heavy reliance on synthetic data generated by Unreal Engine, which provides insufficient priors for real-world visual distributions. Moreover, it exhibits little responsiveness to text-driven semantic viewpoint control. TrajCrafter~\cite{yu2025trajectorycrafter} frequently produces noticeable visual distortions, especially in complex scenes. This is likely because its 3D reconstruction pipeline is less robust under challenging scene geometry, while inconsistencies in the scale of estimated camera parameters further limit its generalization ability. Similar to CamClone, it also shows limited responsiveness to viewpoint instructions. In contrast, although SD-2.0 can respond to semantic viewpoint prompts, it often fails to follow the target camera trajectory. Our method effectively combines both capabilities, producing high-quality videos while accurately following the target camera trajectory and preserving spatio-temporal consistency. Furthermore, the examples on the right side of Figure~\ref{fig:qualitive} demonstrate the strong text-driven semantic viewpoint control of our method, enabling flexible manipulation of shot scale, viewing angle, and first-/third-person perspective.

\subsubsection{Quantitative comparison.}

The quantitative comparison results are summarized in Table~\ref{recam_comparison}. Our method consistently outperforms the baseline approaches across most evaluation metrics, firmly validating the effectiveness and superiority of our proposed framework. 

In terms of camera accuracy, our method outperforms all baselines. Compared with the strongest baseline CamClone~\cite{luo2025camclonemaster}, TARS achieves relative improvements of 15.90\% in rotation accuracy and 41.34\% in translation accuracy. This gain is mainly attributed to our camera grid representation, which enables large-scale self-supervised training to learn accurate camera motion. TrajCrafter~\cite{yu2025trajectorycrafter} directly applies the target camera parameters to the 3D reconstruction of the source video, ignoring scene scale discrepancies and thus producing inaccurate camera motion. SD-2.0~\cite{seedance2026seedance} is designed for multi-reference video generation rather than camera re-shooting, resulting in the weakest performance.

In terms of spatiotemporal consistency, our method achieves the best performance on V-MPGE, demonstrating superior motion alignment with the reference videos. 
Moreover, our method significantly outperforms existing re-shooting baselines on ArcFace, GCR, and LSR, while achieving performance comparable to the commercial model SD2.0. These results indicate that our method effectively preserves identity and scene consistency under novel camera trajectories. We attribute this advantage to the large-scale self-supervised training, which enables the model to learn a rich video prior and maintain the original data distribution during the re-shooting process. Furthermore, our method achieves the highest FSCS among all compared methods, indicating superior semantic consistency in both scene and motion. In contrast, TrajCrafter performs the worst in terms of spatiotemporal consistency, primarily because its 3D reconstruction pipeline struggles to preserve the source video content under large camera motions.

In terms of visual quality, our method achieves the best performance in content expansion plausibility. This improvement is mainly attributed to our self-supervised training framework, which exposes the model to a much broader data distribution. In contrast, TrajCrafter~\cite{yu2025trajectorycrafter} relies on 3D reconstruction to infer unseen regions, resulting in the weakest content expansion capability. Our method achieves visual distortion rates comparable to the commercial model SD-2.0. Notably, the limited robustness of TrajCrafter's reconstruction in complex scenes introduces severe artifacts, leading to the worst distortion performance.

In addition, we report viewpoint control accuracy in Table ~\ref{tab:viewpoint_comparison}. Compared with SD-2.0~\cite{seedance2026seedance}, our method achieves more accurate viewpoint transitions across shot scale, viewing angle, and perspective, demonstrating stronger controllability over viewpoint changes. We attribute this improvement to the strong generalization ability acquired through our large-scale self-supervised training. Notably, our method shows a larger advantage over SD-2.0 in perspective control, primarily because it handles first-/third-person viewpoint transitions for animals more effectively.

\subsection{Ablation Study}

\subsubsection{The Effect of Self-Supervised Learning.}
To evaluate the contribution of self-supervised learning, we remove the self-supervised stage and train the model solely on the limited cross-pair data. As shown in the ``Stage 2'' row of Table~\ref{abaltion_camera} and Table~\ref{tab:viewpoint_ablation}, all quantitative metrics consistently deteriorate. These results indicate that the limited cross-pair data alone are insufficient to learn robust video generation priors. In contrast, large-scale self-supervised learning exposes the model to diverse real-world videos, equipping it with rich priors for camera transitions, appearance, and viewpoint control, thereby providing a strong foundation for the subsequent motion alignment stage.

\subsubsection{The Effect of Cross-Pair Fine-Tuning.}
The ``Stage 1'' results in Table~\ref{abaltion_camera} show that self-supervised learning alone establishes strong generative priors but cannot ensure motion synchronization due to the temporal misalignment of the training pairs. Adding the second-stage cross-pair fine-tuning consistently improves all metrics, especially spatio-temporal consistency, demonstrating that a small amount of strictly paired data is sufficient to align subject motion while preserving the generative priors learned during self-supervised training. The viewpoint control results in Table~\ref{tab:viewpoint_ablation} further verify that this fine-tuning does not degrade the learned viewpoint control capability.

\subsubsection{Effect of timestep-aware signal injection.}
To further validate our observation, we inject the source video only at either the high-noise or the remaining timesteps at inference time. As shown in the ``High'' and ``Mid\&Low'' rows of Table~\ref{abaltion_camera}, source injection at high-noise timesteps mainly affects low-frequency attributes, including camera motion (R-Prec and T-Prec) and motion consistency (V-MPGE), whereas injection at the remaining timesteps primarily influences high-frequency appearance. These results verify the distinct learning of different diffusion timesteps and further justify our timestep-aware training strategy.

\section{Conclusion}
In this paper, we present TARS, the first unified framework for controllable video re-shooting with text-driven semantic viewpoint control. We observe that different diffusion timesteps play distinct roles in video generation, motivating our timestep-aware two-stage training strategy. Specifically, large-scale self-supervised learning establishes rich generative priors, while limited cross-pair fine-tuning in the high-noise regime enables accurate motion alignment. Extensive experiments demonstrate that TARS achieves state-of-the-art performance in camera accuracy, semantic viewpoint control, spatio-temporal consistency, and visual quality.

\bibliography{aaai2027}

\end{document}